%% file: acl_latex.tex
\pdfoutput=1

\documentclass[11pt]{article}

\usepackage[final]{acl}

\usepackage{times}
\usepackage{latexsym}

\usepackage[T1]{fontenc}

\usepackage[utf8]{inputenc}

\usepackage{microtype}

\usepackage{inconsolata}

\usepackage{graphicx}

\usepackage{booktabs}
\usepackage{multirow}
\usepackage{listings}
\usepackage{longtable}

\newcommand{\dataname}{\textsc{PeopleJoin}}
\newcommand{\asyncfw}[0]{\dataname}
\newcommand{\dataspider}{\textsc{PeopleJoin-QA}}
\newcommand{\datanews}{\textsc{PeopleJoin-DocCreation}}

\newcommand{\reactagent}{\textsl{Reactive}}
\newcommand{\noreflectionagent}{\textsl{Reactive-NoRef}}

\newcommand{\messageall}{\textsl{MessageAllOnce}}
\newcommand{\messagenone}{\textsl{MessageNone}}
\newcommand{\idealagent}{\textsl{IdealAgent}}

\definecolor{keywordcolor}{rgb}{0.13, 0.29, 0.53}  
\definecolor{customkeywordcolor}{rgb}{0.75, 0.0, 0.75}  
\definecolor{stringcolor}{rgb}{0.71, 0.21, 0.01}   
\definecolor{commentcolor}{rgb}{0.0, 0.5, 0.0}     
\definecolor{bgcolor}{rgb}{0.95, 0.95, 0.95}       

\lstdefinelanguage{Markdown}{
    keywords=[1]{\#}, 
    keywordstyle=\color{keywordcolor},  
    comment=[l]{\# },                    
    commentstyle=\color{commentcolor},  
    stringstyle=\color{stringcolor},     
    morestring=[b]`,                     
}

\title{LM Agents for Coordinating Multi-User Information Gathering}

\author{
 \textbf{Harsh Jhamtani},
 \textbf{Jacob Andreas},
 \textbf{Benjamin Van Durme}
\\
 Microsoft
\\
 \small{
 \{hjhamtani,jaandrea,ben.vandurme\}@microsoft.com
 }
}

\begin{document}
\maketitle
\input{files/0.abstract}

\input{files/1.intro}
\input{files/2.data}

\input{files/3.approach}

\input{files/5.2.0.results.table} 
\input{files/5.1.setup}

\input{files/5.2.results}

\input{files/5.3.humaneval}

\input{files/8.related}

\input{files/9.conclusion}

\bibliography{acl_latext}

\newpage
\appendix
\section*{Appendix}
\input{appendix/10.4.approach.appendix}

\input{appendix/10.6.expsetup.appendix}

\input{appendix/10.0.data.appenidix}

\input{appendix/10.4.1}

\input{appendix/10.6.1}

\end{document}

%% file: files/0.abstract.tex
\begin{abstract}

This paper introduces \asyncfw{}, a  benchmark for evaluating LM-mediated collaborative problem solving. Given a user request, \asyncfw{} agents must identify teammates who might be able to assist, converse with these teammates to gather information, and finally compile a useful answer or summary for the original user. \asyncfw{} comprises two evaluation domains: \dataspider{}, focused on questions about tabular data, and \datanews{}, focused on document creation tasks. The two domains are adapted from existing NLP benchmarks for database question answering and multi-document summarization; here, however, the information needed to complete these tasks is distributed across synthetic ``organizations'' of 2--20 users, simulating natural multi-user collaboration scenarios.
We implemented several popular LM agent architectures, evaluating their accuracy and efficiency at completing tasks, and highlight new research questions that can be studied using \asyncfw{}.
\footnote{Code and data can be found at \url{https://github.com/microsoft/peoplejoin/}}

\end{abstract}

%% file: files/1.intro.tex
\section{Introduction} 

\begin{figure*}
    \centering
    \includegraphics[width=.8\linewidth,trim=0.3in 1.4in 0.3in 1.4in]{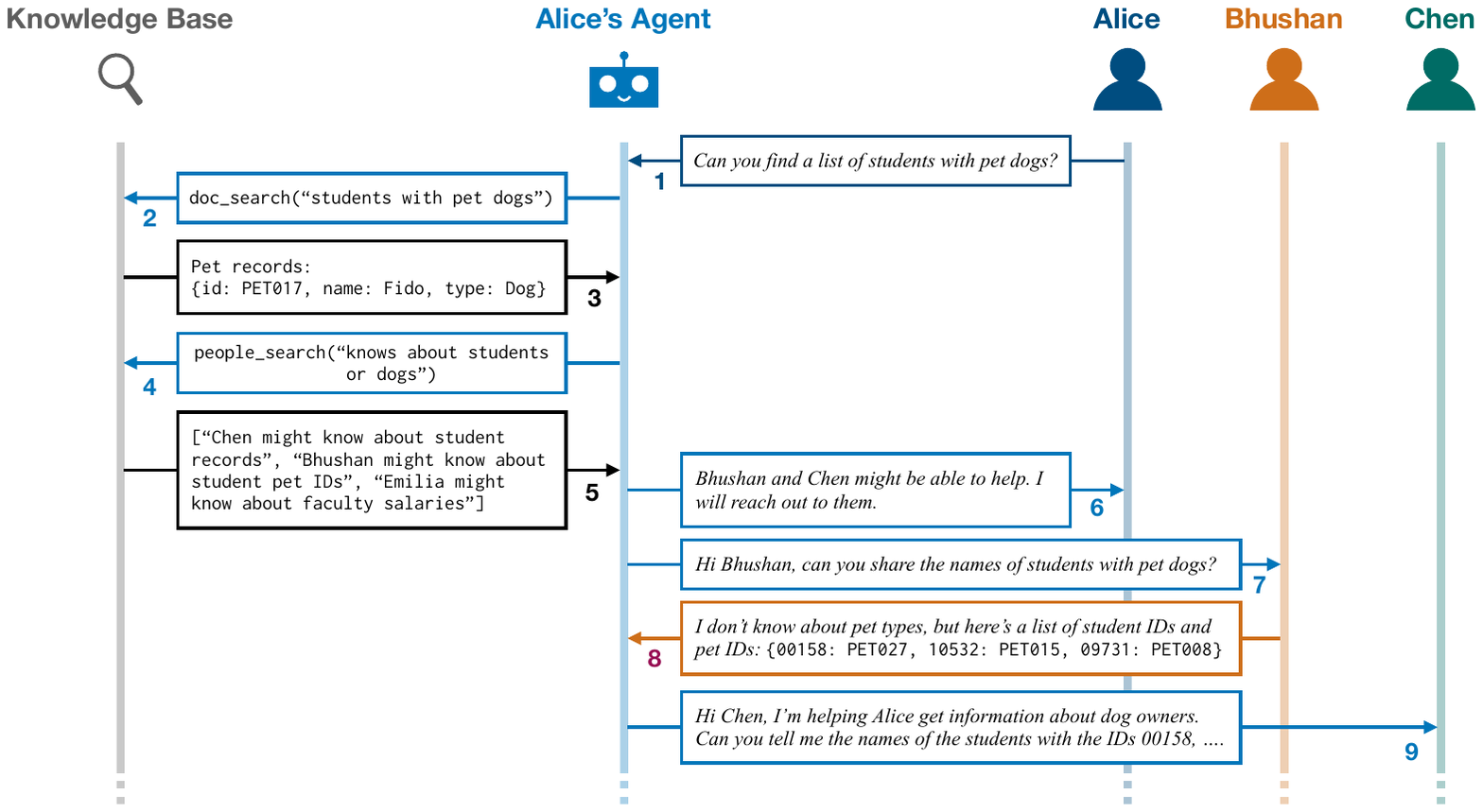}
    \caption{A sequence diagram illustrating a conversation in \asyncfw{} framework, where Alice issues a request to her agent. Documents available to Alice's agent are insufficient to answer the user request. The agent uses a people search tool, after which it decides what subset of people to contact, in which order, what questions to pose, etc. The temporal ordering of tool calls and message exchanges is denoted by \#i. }
    \label{fig:intro}
\end{figure*}

In today's fast-paced and interconnected world, effective collaboration is essential for achieving complex tasks and making informed decisions \cite{papachristou2023leveraging,gemp2024states}. 
Many decision-making, content creation, and information-gathering tasks require collecting information from multiple people. For example, preparing a list of interns across teams in an organization by reaching out to the leader of each team; preparing a newsletter for project updates might necessitate coordinating with multiple contributors; identifying a suitable time to meet might require several rounds of negotiations \cite{lin2024decision}. Identifying what information is available, judiciously determining who to contact, asking precise questions, and compiling research results can be a challenging and time-consuming process---especially when real-time interaction between team members is difficult to coordinate.

At the same time, recent large language models (LLMs), such as GPT-4 \cite{openai2023gpt4}, Phi-3 \cite{abdin2024phi}, LLaMa \cite{touvron2023llama}, and Gemini \cite{team2023gemini}, are becoming a crucial building block in developing automated agents that can assist human users with complex tasks
\cite{xi2023rise,wang2024survey,butler2023microsoft}.
These tasks include chat applications for assisting individual users with searching and summarizing information (such as in Microsoft Copilot Chat\footnote{\url{https://copilot.microsoft.com/}}), and even supporting these users in workplace decision-making \citep{butler2023microsoft,kim2024leveraginglargelanguagemodels}.
Could these agents be extended to improve collaboration among multiple users?

In this paper, we introduce
\textbf{\asyncfw{}},
an evaluation framework for studying effectiveness of LLM-powered agents to assist with multi-user collaboration tasks.
Each \asyncfw{} task takes place within a fictitious organization with 2--20 employees, some of whom possess a collection of documents necessary to solve some task.
One of the users (the \emph{initiating user}) communicates the task to an \emph{agent} (Fig.~\ref{fig:intro}).
Agents have direct access to the initiating user's documents, and can engage in conversations with other users to gather relevant information. 
They must rely on limited descriptions of other users, and potentially previous interactions, to determine who to contact for a given task.
\asyncfw{} comprises two families of tasks: \textbf{\dataspider{}} and \textbf{\datanews{}}, derived from the \textsc{spider} \citep{yu2018spider} and \textsc{multinews} \citep{fabbri2019multi} datasets respectively. It evaluates 
agents' ability to answer questions involving complex relational reasoning and document summarization.

Our initial benchmark release also includes reference agent implementations based on popular
prompting and orchestration strategies, and a suite of evaluation metrics.
We report evaluation results using Phi-3-medium \cite{abdin2024phi}, GPT-4-turbo and GPT-4o \cite{openai2023gpt4} language models  to implement these agents. Our results indicate that LM-powered agents can struggle to coordinate with multiple users to correctly address information seeking and document authoring requests. Major research questions remain around how to optimally determine which people to contact and in what order, how to ask high-quality questions, and how to learn and adapt to the structure of an organization. 
\asyncfw{} thus provide a test-bed for building AI-driven systems that can enhance human collaboration, and will also enable future work on learning from interaction, distributing tasks equitably, and maintaining user privacy in such agentic systems.

%% file: files/2.data.tex
\section{Challenges in Effectively Steering Multi-User Information Gathering}

The problem of answering user queries by synthesizing information distributed across heterogeneous data sources is most often studied through the lens of database systems \cite{zaniolo1997advanced}.
Work on query optimization and federated databases \cite{sheth1990federated} has sought to address the 
specific question of how to efficiently answer structured queries without access to a centralized knowledge store. 
The problem we study in \asyncfw{} may be viewed as a generalization of this task to the setting where the relevant information is possessed by \emph{people}, not structured knowledge bases, and must be obtained via conversation rather than structured queries.
An agent to help a user with such requests must address several challenges:

\begin{itemize}

    \item \textbf{Information fragmentation:} 
    In a typical organization, information is often siloed across multiple users, because of differing roles and responsibilities. 
    Some requests may require gathering information from multiple people.

    \item \textbf{Partial observability:} 
    To gather this information, it is often necessary to first determine which collaborators hold relevant information, under incomplete and potentially imprecise information of what information each collaborator might have. Agents for collaborative decision-making might have to engage in multi-turn conversations with various users, refining and adapting requests as needed.
    
    \item \textbf{Communication costs:} Requests for information require human effort to process and answer; effective collaboration requires \emph{efficient} communication: 
    effective agents
    should judiciously send information requests to other collaborators, 
    and avoid asking questions that are likely to be unanswerable.
    
    \item  \textbf{Complex reasoning and planning:} Efficient communication requires reasoning: establishing what information is available in accessible documents, dynamically predicting which collaborators are likely to have relevant information for specific questions, identifying the best order in which to ask these questions, and re-planning based on collaborators' responses.

\end{itemize}
Below, we present a benchmark for evaluating these skills.

\section{Data}
\label{sec:data}

Each \asyncfw{} domain comprises a set of \textbf{organizations}. Each organization contains a set of \textbf{collaborators}, and each \textbf{collaborator} has privileged access to a set of \textbf{documents}. The benchmark provides LLM-based simulators for each collaborator, a search interface that can be used to find collaborators,
and a messaging interface that can be used to ask collaborators about their documents.
Then an \textbf{agent} must take as input a \textbf{query} from one collaborator, use the search and messaging interfaces to interact with other collaborators, and finally return an \textbf{answer} to the originator.

Drawing analogies between multi-user collaboration tasks and existing multi-\emph{data-source} tasks commonly studied in NLP, we develop \asyncfw{} by re-purposing existing high-quality resources for database question answering (to produce \dataspider{}) and multi-document summarization (to produce \datanews{}).

\begin{figure*}
    \centering
    \includegraphics[width=.8\linewidth,trim=0.1in 3.3in 0.1in 1.2in]{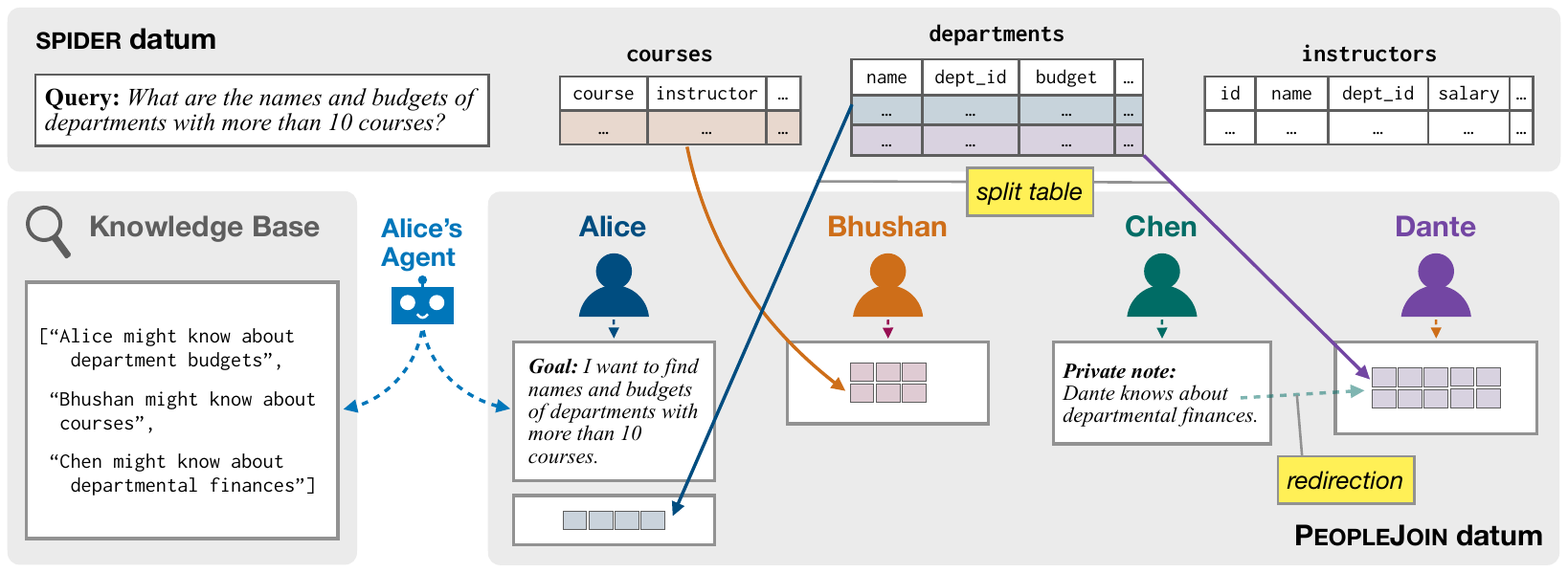}
    \caption{Illustration of a transformation of a Spider datum into \dataspider{}.}
    \label{fig:data-transformation}
\end{figure*}

\subsection{\dataspider{}}
The \dataspider{} dataset 
evaluates LM agents' abilities to answer questions by aggregating information from multiple collaborators.
We construct it by re-purposing \textsc{spider} \cite{yu2018spider}, a text-to-SQL benchmark. We transform \textsc{spider} into a multi-user information gathering task by recasting \textsc{spider} tables as ``documents'', distributed among several users, and interpreting
\textsc{spider} questions
as queries from an initiating user to an AI agent. 
In this scenario, answering questions requires identifying which users possess the relevant pieces of information (similar to selecting tables in a database), and then engaging in multi-turn conversations with these users to ask targeted questions (akin to constructing joins between tables). 

\textsc{spider} consists of a set of 200 databases, with a total of over 10K questions. 
Each database in \textsc{spider} is transformed into an ``organization'' containing a set of $2$--$20$ distinct users, each with access to a distinct set of documents.

\paragraph{Documents}
Each table in a \textsc{spider} database is converted to one or more documents.\footnote{Represented as a sequence of \textsc{json} objects, one per row.} We additionally apply the following transformations to elicit a diverse set of information-gathering behaviors:

\begin{enumerate}
     \item \textbf{Split Documents:}  One of the randomly selected tables is split into two parts (each containing half the rows). This simulates a scenario in which information about a given topic is distributed across multiple individuals. 
     For instance, in Fig.~\ref{fig:data-transformation}, the information in the table \texttt{department} is split between Alice and Dante. 
    \item \textbf{Redirection:} We construct scenarios in which a (``redirecting'') user does not have direct access to some information (e.g.\ Chen in Fig.~\ref{fig:data-transformation}), but does have knowledge of which other (``target'') user might have this information (Dante in Fig \ref{fig:data-transformation}). To answer questions about these tables, agents cannot always contact knowledgeable users directly, and must navigate organizational knowledge hierarchies to find them.
    Information about other users is available to the redirecting user as an additional document.
    \item \textbf{Missing Information:} 
    In each database,
    we omit a randomly selected table, making a subset of the queries associated with that organization \textbf{unanswerable}, simulating a scenario in which required information is simply not present~\cite{levy-etal-2017-zero, rajpurkar2018know} in the organization. 
   
\end{enumerate}

\noindent{}
In \dataspider{}, each user is allocated one document, and no two users have access to the same document. 
After we have assigned each organization member a set of documents, we populate the collaborator search interface with hints about what information they might have access to (e.g.\ \emph{Chen likely has information about teacher salaries}).
We begin by constructing templated descriptions specifying the table name and names of columns, then use GPT-4 to convert these to simpler English statements using a few-shot prompting setup.
These transformations by design sometimes result in imprecise or incomplete descriptions, simulating the challenges of selecting a good subset of people to contact under limited information.
For example, \emph{Chen might know about student demographics} fails to specify what specific demographic information is there, and how it is associated with students (e.g., using a student ID, name, or other identifier).
For redirections as described above, these descriptions state that the redirecting user has the information that is in fact possessed by the target user.

\paragraph{Task and Evaluation}

Each organization is associated with multiple problem instances, one for each question in the underlying \textsc{spider} dataset.
For example, in Fig \ref{fig:data-transformation}, the task issued by Alice  to their agent is \emph{What are the name and budgets of the department...}, which must then be answered by reasoning about the contents of both Alice's documents and the other users'. Ground-truth answers are derived from the underlying \textsc{Spider} annotations, except in the case of un-answerable queries. Our primary evaluation metric measures the accuracy with which agents can recover ground-truth answers and identify unanswerable questions; secondary evaluations measure whether the right users were contacted, and the efficiency (in terms of messages sent) with which agents identify these users.

\paragraph{Statistics}
Though the \textsc{Spider} dataset has several thousands of questions paired with databases, we restrict to 500 test tasks to enable efficient evaluation.\footnote{Our code release makes it possible to generate additional organizations for training and evaluation.} A typical task requires agent to interact with 0--5 people (excluding the initiating user) to arrive at the answer (mean of 1.54 with a variance of 1.12).
9\% of test instances in \dataspider{} are \emph{unanswerable},  22\% of the test instances require an agent to handle a \emph{redirection} to arrive at correct answer, while 25\% of the test instances  require an agent to handle a \emph{document split} between multiple people to answer the user question correctly. 
Note that a data instance could belong to more than one category (for example, a task might require access to \emph{split documents} as well as access to information from another document that needs to be accessed through \emph{redirection}).

\subsection{\datanews{}}

The \datanews{} task evaluates agents not on structured QA, but instead on more open-ended document creation tasks. We derive it from  \textsc{MultiNews}  \cite{fabbri2019multi}, a multi-document summarization dataset consisting of sets of news articles on a related topic and single summaries that aggregate information across the articles.
We
distribute source news articles across multiple users, and require agents to gather these documents (or excerpts from them) and combine them into a target summary. 

\paragraph{Task and Evaluation}

As in \dataspider{}, each organization is derived from underlying \textsc{MultiNews} problem instances. Here, however, \emph{multiple} problem instances are combined into a single organization: some users have articles on one subject, some users have articles about multiple subjects, and some may have no articles at all. Each organization possesses information about $3$ topics, and contains $1$--$7$ users, with documents randomly partitioned across users.

Also as in \dataspider{}, we create user descriptions for collaborator search  by presenting user documents to GPT-4 and querying it for a list of keywords that the user is knowledgeable about (e.g.\ \emph{governor election, GOP, health care}).

\paragraph{Statistics}

Because of the relatively large size of the documents that must be exchanged to complete these tasks, we construct 200 test instances distributed across 67 organizations.
Summaries are derived from an average of 2.7 documents (variance of 1.1), which must be located within organizations with an average of 5.1 users (variance of 4.5) and 6.4 total documents (variance of 4.1) or 1.25 documents per user.

%% file: files/3.approach.tex
\section{Baseline Agent Architectures}

To demonstrate the usefulness of \asyncfw{} as a research platform, we develop and evaluate a reference  
LM-powered agent implementation to perform tasks by coordinating interactions, retrieving relevant information, and posing targeted queries to other organization members. 
We consider an event-based reactive agent, which is triggered by user actions: upon getting a message from any organization member, the agent follows ReAct-style prompting loop \cite{DBLP:conf/iclr/YaoZYDSN023}, taking actions, making observations, and performing reflection, until it decides to pause and wait for a next event, or terminate the session.

\subsection{Actions}

The agent can perform a few types of actions.
\textbf{Document Retrieval}:
agents have access to documents accessible to the initiating user, by invoking a function \texttt{search\_documents(query: str)}. Documents are indexed using a standard BM25 index, and the tool call returns a fixed number (upto 3) of documents with the highest matching score. 
\textbf{People Retrieval}:
agents can search through a repository of employee profiles and knowledge areas, by invoking a function \texttt{search\_relevant\_people(query: str)}. 
However, these expertise profiles may be outdated or imprecise, requiring the agent to navigate uncertainty while coordinating queries. As in document retrieval, descriptions are retrieved using a standard BM25 index. A fixed number (up to 10) of highest-scoring results are returned.
\textbf{Sending Messages:} 
the agent is capable of exchanging messages with any person in the organization. 
\textbf{Person Resolution}:
the agent can resolve a person name to get their user ids, to be used to send messages to them.
\textbf{Turn and Session Completion}: agent can mark the current turn or the entire session as completed. 

Signatures of Python functions corresponding to the allowed actions are provided in the prompt. See Appendix \ref{sec:action_descriptions} for the full set of action descriptions. 

\subsection{Observations and Reflection}

After each action is taken, the agent receives a textual observation.
These include retrieved documents or descriptions of collaborators. 
As is typical in LLM-based agent architectures, these observations are simply appended to the agent's prompt. Before invoking additional actions, the agent may perform \emph{reflection} actions, corresponding to text-based (``scratchpad'' or ``chain-of-thought'') reasoning about its future plans. Our agent represents reflection as tool calls that return no value but remain in the agent's prompt at future timesteps.


\subsection{Prompt Structure}

The prompt has 3 parts: action descriptions (outlined above); exemplars; and interaction history.

\textbf{Exemplars:} In each domain, we manually annotated four exemplars (See Appendix \ref{sec:exemplars} for a full exemplar) with events, actions, and observations. The exemplars are designed to reflect all relevant phenomena in the domain in question, such as dealing with fragmented information, handling unanswerable questions, and managing redirection. 

\textbf{Interaction History:} An event (receiving a message from an employee) triggers LLM into a loop of action prediction, observation, and reflection, till an end of turn or session is predicted. 
Actions are executed immediately after they are predicted;
events, action, and observation are incrementally appended in the prompt in the order in which they occur (see Appendix~\ref{sec:appendix-approach}).


%% file: files/5.2.0.results.table.tex
\begin{table*}[t!]
\footnotesize
\centering
\begin{tabular}{lcccccc}
\toprule
\multirow{2}{*}{\textbf{Method}} & \multicolumn{1}{c}{\textbf{Outcome}} & \multicolumn{3}{c}{\textbf{Task Efficiency}} & \multicolumn{2}{c}{\textbf{Info Source}} \\
\cmidrule(r{4pt}){2-2}
\cmidrule{3-5}
\cmidrule(l{4pt}){6-7}
      & \textbf{Match} $\uparrow$ & \textbf{MsgCnt} $\downarrow$ & \textbf{MsgSize} $\downarrow$ & \textbf{\#People} $\downarrow$ & \textbf{P-Prec}$\uparrow$ & \textbf{P-Rec}$\uparrow$ \\
\midrule
LLM: \texttt{gpt-4-turbo}    &           &   &          &   &          & \\ 
\textbf{\reactagent{}}      &  \bf 54.8  &        9.0   & 193  & 1.5   &  0.61  &  0.89         \\
\textbf{\noreflectionagent{}}     &  48.0  &   9.2        &  187  &   1.5 &  0.55    &  0.82      \\

\midrule
LLM: \texttt{gpt-4o}     &    &   &          & & & \\ 
\textbf{\reactagent{}}         &  48.7         & 9.7  &    179    & 1.2  &    0.60      &   0.83   \\
\textbf{\noreflectionagent{}}   &     40.4      &  10.4  &  209   & 2.0  &    0.52      &  0.78      \\ 
\midrule
LLM: \texttt{phi-3-medium}    &           &   &          &   &          &  \\ 
\textbf{\reactagent{}}    &  24.4         & 6.7  &    122      &  1.0  &       0.23   &  0.52      \\
\textbf{\noreflectionagent{}}               &  20.0 &   16.3 & 295  & 1.7     &  0.39        &   0.62     \\
\bottomrule
\end{tabular}
\caption{Results on \dataspider{}.
}
\label{tab:results-spider}
\end{table*}

\begin{table*}[t!]
\footnotesize
\centering
\resizebox{0.95\textwidth}{!}{
\begin{tabular}{lccccccc}
\toprule
\multirow{2}{*}{\textbf{Method}} & \multicolumn{2}{c}{\textbf{Outcome}} & \multicolumn{3}{c}{\textbf{Task Efficiency}} & \multicolumn{2}{c}{\textbf{Info Source}} \\
\cmidrule(r{4pt}){2-3}
\cmidrule{4-6}
\cmidrule(l{4pt}){7-8}
      & \textbf{Rouge} $\uparrow$  & \textbf{G-Eval} $\uparrow$ & \textbf{MsgCnt} $\downarrow$ & \textbf{MsgSize} $\downarrow$ & \textbf{\#People} $\downarrow$ & \textbf{P-Prec}$\uparrow$ & \textbf{P-Rec}$\uparrow$ \\
\midrule
LLM: \texttt{gpt-4-turbo} \\
\textbf{\reactagent{}}  & 16.3 & 4.00 / 4.16 / 4.07  & 12.6 & 1330 & 1.5 & 0.99 & 0.88 \\
\textbf{\noreflectionagent{}}   & \bf 16.5 & 4.20 / 4.33 / 4.14  & 12.4 & 1281 & 1.5 & 0.97 & 0.87 \\
\midrule
LLM: \texttt{gpt-4o} \\
\textbf{\reactagent{}}  & 12.2 & 2.99 / 3.33 / 3.00 & 9.9 & 1180 & 1.4 & 0.95 & 0.80 \\
\textbf{\noreflectionagent{}}   & 12.6 & 3.15 / 3.42 / 2.65 & 10.9 & 1268 & 1.7 & 0.90 & 0.90 \\
\midrule
LLM: \texttt{phi-3-medium} &  \\ 
\textbf{\reactagent{}} &  11.5 &  2.84 / 3.31 / 2.81 &  11.0 &   996       & 1.7  &     0.66     & 0.69 \\

\textbf{\noreflectionagent{}}  &  11.3 & 2.71 / 2.64 / 3.20 & 11.3 & 948 & 1.7 & 0.65 & 0.67 \\
\bottomrule
\end{tabular}
}
\caption{Results on \datanews{}. G-Eval consists of three scores (Relevance/Consistency/Coherence).
}
\label{tab:results-news}
\end{table*}

%% file: files/5.1.setup.tex
\section{Evaluation}

\asyncfw{} provides metrics for evaluating the efficiency and correctness of user interactions. 

\subsection{Outcome Metrics}

The most important measure of an agent's effectiveness is its ability to provide the correct response to the user's query. We characterize correctness in different ways for the  domains within \asyncfw{}.

\paragraph{Answer match:} For \dataspider{}, we prompt an LLM-based evaluator to compare the agent's final response to the reference answer and output a score in \{0,50,100\}, where a score of 100 refers to a perfectly matched score (all the expected information was present), a score of 50 refers to a partial match (for example, if only few of the expected list of items were correctly provided), while a score of 0 refers to incorrect results (for example, if the agent claimed it could not find the requested information but gold answer suggests otherwise).  
The score is predicted by an LLM (\texttt{gpt-4-turbo}), conditioned upon the agent response to the initiating user and the expected gold answer, certain prompt instructions and three examples. More details are available in the Appendix \ref{appendix-match-score}. 

\paragraph{\textsc{rouge and G-Eval}} 
For the \datanews{} task, we require agents to output a final summary enclosed by special delimiter tokens, then report the \textsc{rouge-L} score \citep{lin2004rouge} of this summary relative to the reference summary. If the agent produces no summary, it obtains a score of 0; if it produces multiple summaries on different turns,
 we score only the final one.
 We also report \textsc{G-Eval} scores \citep{DBLP:conf/emnlp/LiuIXWXZ23}, a set of automated metrics that evaluate the relevance, consistency, and coherence of a summary using an LM with access to 
source documents.
 G-Eval has been found to correlate highly with human summarization ratings \cite{DBLP:conf/acl/SongSSCM24}.

\subsection{Efficiency Metrics}

An effective agent should not only produce correct answers, but do so while minimizing effort from collaborators. We quantify this using three metrics.
\textbf{Message count (Msg)}: measures the total number of messages exchanged during the task.
\textbf{Message size (MsgSize)}: message count alone does not penalize requests requiring lengthy responses from collaborators, so we additionally report the total number of words exchanged (tokenized using the NLTK \cite{bird2009natural} word tokenizer). 
\textbf{People contacted (\#People)}: the count of people that the agent exchanged messages with (including the initiating user), averaged across the test set.

\subsection{Information Source Metrics}

In both \dataspider{} and \datanews{}, the gold set of documents required to answer a task correctly are known, which also allows us to infer the \emph{optimal set of people} an agent must contact to arrive at the correct outcome. We collect the set of distinct users contacted by the agent, then compute the precision (\textbf{P-Prec}) and recall (\textbf{P-Rec}) relative to the ground-truth people set, averaged across queries.

%% file: files/5.2.results.tex
\section{Experiments}

The \asyncfw{} framework includes user simulators that represent collaborators within an organization, along with scaffolding code that enables an agent to search through the initiating user's documents and identify and contact relevant collaborators.
All experiments use a \texttt{gpt-4-turbo} model \citep{openai2023gpt4}, prompted with each collaborator's description and document collection, to implement these simulators (full prompt in Appendix \ref{appendix-user-sim}). We then evaluate our reference agent architecture using the metrics described above.


We compare several alternative implementations of this reference architecture, including variations in task orchestration and planning strategies. 
\textbf{\reactagent{}} is the full agent architecture
\cite{DBLP:conf/iclr/YaoZYDSN023}, and
\textbf{\noreflectionagent{}} is a variant of this architecture which performs no reflection actions.
We compare \texttt{gpt-4-turbo} \cite{openai2023gpt4}, \texttt{gpt-4o} \cite{openai2023gpt4}, and \texttt{phi-3-medium} \cite{abdin2024phi} as LLMs.
We use greedy decoding.

\subsection{Results on \dataspider{}} 
The max score on Match metric across all methods is only $54.8$ (Table \ref{tab:results-spider}), achieved by \reactagent{} when used with \texttt{gpt-4-turbo}, demonstrating the overall challenging setup. 
Moreover, for the same configuration, P-Prec and P-Rec scores are $0.61$ and $0.89$ respectively, demonstrating scope of further improvement in optimal selection of people to contact.
Comparing LLM choices for \reactagent{}, \texttt{gpt-4-turbo} performed better than \texttt{gpt-4o}, while \texttt{phi-3-medium} is generally worse on Match and information source selection.
Finally, \reactagent{} 
 generally performs similar or better than \noreflectionagent{} across LLMs on Match, efficiency, and optimal selection of information sources, demonstrating the usefulness of a \emph{reflection} step. \\

 \noindent \textbf{Additional Comparisons:}
To put these results in perspective, we additionally compare with following techniques:\\
\begin{table}[b!]
\footnotesize
\centering
\begin{tabular}{l@{\hspace{5pt}}cc@{\hspace{5pt}}cc@{\hspace{5pt}}c}
\toprule
& \textbf{Match} $\uparrow$ & \textbf{MsgCnt} $\downarrow$ & \textbf{P-Prec} $\uparrow$ \\
\midrule
\reactagent{} & 54.8 & 9.0 & 0.61 \\
\messageall{} & 34.6 & 11.4 & 0.37 \\
\messagenone{} & 19.2 & 4.1 & N/A \\
\idealagent{} & 100 & 7.0 & 1.0 \\
\bottomrule
\end{tabular}
\caption{Additional Comparisons (using \texttt{gpt-4-turbo})}
\label{tab:additional-comparisons}
\end{table}
\noindent (1) \textbf{\messageall{}}, an agent that is encouraged (through prompt instructions and exemplars) to message each person in the organization exactly once, with the same question the user asked.  \messageall{} results highlight the importance of judiciously choosing who to contact (MsgCnt of 11.4 compared to 9.0 for \reactagent{}), framing the correct questions and engaging in multi-turn conversations with collaborators when needed (Match score is much lower than that of \reactagent{}). \\
\noindent (2) \textbf{\messagenone{},} an agent that attempts to complete the task with the user's documents alone (i.e. without contacting any collaborator).
\messagenone{} results provide a baseline performance when no collaborator is contacted. \\
\noindent (3) \textbf{\idealagent{}}, which is defined as the one that always gets the correct answers by contacting the optimal set of relevant collaborators, formulating perfect questions, etc.
will get a Match score of 100, \#People count of 1.5 (equals count of the optimal set of people to contact), and MsgCnt of 7. \\

\noindent \textbf{Analysis:}
We analyzed Match scores on subsets of \dataspider{} for \reactagent{} with \texttt{gpt-4-turbo}: 
(1) \emph{Document Split}: 50.0; 
(2) \emph{Redirection}: 38.0;
(3) \emph{Unanswerable}: 87.5. 
The results demonstrate that \reactagent{} does particularly well in identifying unanswerable questions, but struggles with information fragmentation and knowledge hierarchies required to correctly handle the redirection category. 

We include a few qualitative examples in Appendix \ref{appendix-qual}. Additionally, we analyzed 40 random examples with imperfect Match scores in PeopleJoin-QA when using \reactagent{} and the most common failure modes were: (1) Failing to contact all  relevant users and arriving at an incorrect answer [$30\%$ of cases]. 
(2) Poorly worded or overly-specific queries from the agent causing other users to conclude that they didn’t have relevant information [$25\%$ of cases]. For example, the Listing 9 in Appendix B.3. (3) Failing to reach out to all the relevant people and telling the user it couldn’t get all the information [$20\%$ of cases]. 
(4) Orchestration errors, such as not predicting tools for people or document search [$10\%$] (Listing 8 is an example). 

\subsection{Results on \datanews{}}
On \datanews{}, among the LLM choices, \texttt{gpt-4-turbo} performs better than \texttt{gpt-4o}, which in turn performs better than \texttt{phi-3} (Table \ref{tab:results-news}). In contrast to results in \dataspider{}, \reactagent{} and \noreflectionagent{} variants perform similar, suggesting no usefulness of the reflection step in the document creation task. 
On this task, an \idealagent{} should obtain G-Eval scores of 5, MsgCnt of 6.3, MsgSize of 1592, and \#People of 1.7.
These results indicate that the document creation task is also challenging, with significant scope for improvement in output quality and communicative efficiency.
\\

\noindent \textbf{Analysis:} Here, the most common failure modes (in 40 analyzed examples) were (1) failing to ask follow-up questions in cases where one user had multiple documents on a given topic [38\% of cases], (2) poorly worded or overly-specific queries, causing other users to conclude that they didn’t have relevant documents [24\%], and (3) orchestration failures in which the agent was distracted by a user comment and ended the conversation early or stopped pursuing the original goal [38\%].

%% file: files/5.3.humaneval.tex
\subsection{Case Study with Human Participants}

The experiments discussed above rely on simulated users. To complement this, we conducted a human evaluation study in which real users took on the roles of certain collaborators in the experiment. The goal of this study was to assess whether the agents perform the task with similar efficacy when interacting with human users compared to a fully simulated environment.
Like simulated users, human participants (Appendix \ref{appendix-humaneval})  had access to the documents associated with their assigned personas. Messages from the agent indicated that they were generated by an automated system. While participants were free to respond as they saw fit, they were instructed to engage as respectful colleagues within a business setting.

The study was conducted on 100 randomly selected examples from the \dataspider{} dataset. In each instance, one collaborator role was played by a human participant. To ensure meaningful interaction, rather than selecting personas randomly— which could result in cases where the human collaborator was not contacted by the agent— we specifically picked the human collaborator to be among the gold set of individuals the agent needed to contact for the test example in question.

Table \ref{tab:human-results-spider} presents results, comparing performance metrics between human-in-the-loop interactions and the fully simulated setup, when using \reactagent{} with \texttt{gpt-4-turbo}.
Human collaborators provided slightly longer responses and asked more clarification questions than simulated collaborators, leading to a higher number of messages from the agent as well. 
We also observed slightly better average Match score with human users compared to full simulation. But together, these results suggest the simulated setup produces qualitatively similar dialogs and outcomes to human interactions. 

\begin{table}[t!]
\footnotesize
\centering
\begin{tabular}{l@{\hspace{5pt}}cc@{\hspace{5pt}}cc@{\hspace{5pt}}c}
\toprule
& \textbf{Match} $\uparrow$ & \textbf{MsgCnt} $\downarrow$ & \textbf{MsgSize} $\downarrow$ \\
\midrule
Human Participant & 50 & 10.0  & 198 \\
Simulation & 44 & 9.3 & 187 \\
\bottomrule
\end{tabular}
\caption{Human Evaluation Case Study}
\label{tab:human-results-spider}
\end{table}

%% file: files/8.related.tex
\section{Related Work}

\textbf{AI-mediated collaboration and negotiations:}
Recent research in human-AI collaboration has explored various strategies for facilitating decision-making and negotiations among multiple users. \citet{lin2024decision} examines how AI assistants can assist humans through natural language interactions to make complex decisions, such as planning a multi-city itinerary or negotiating travel arrangements among friends. 
\citet{gemp2024states} focus on how game-theoretic approaches that can guide LLMs in tasks like meeting scheduling and resource allocation. 
Past work \cite{papachristou2023leveraging} has also explored the role of LLMs in facilitating group decisions, such as selecting a meeting time or venue, where LLM agents analyze individual preferences from conversations. 
In contrast, \asyncfw{} focuses on LM agents for coordinating multi-user information gathering.

\textbf{Multi-hop reasoning and task decomposition:}
In our setup, an agent needs to compile information from multiple sources, a theme shared with prior work in multi-hop QA \cite{welbl2018constructing, yang2018hotpotqa} and multi-document summarization \cite{liu2018generating,fabbri2019multi}. 
Past work on solving complex tasks by decomposing them (via prompting) into simpler sub-tasks \cite{wolfson2020break, khotdecomposed, jhamtani2023natural} is also relevant.
Compared to such past work, our setup requires additional steps of finding the relevant users, posing apt questions, compiling the gathered information, and doing so with minimum communication overhead possible.

%% file: files/9.conclusion.tex
\section{Conclusions and Future Directions}

\asyncfw{} is  a new benchmark designed to evaluate the role of language model (LM) agents in facilitating collaborative information gathering within multi-user environments. 
It comprises two domains, \dataspider{} and \datanews{}, which challenge LM  agents to handle tasks related to question-answering and document creation. Experiments with popular LM agent architectures revealed both their potential and limitations in accurately and efficiently completing complex collaborative tasks. 

Future work could consider AI agents that learn over time from interactions for improving their performance over time \cite{lewis1998designing}. By analyzing past conversations, they can improve information source selection and communication strategies, making future interactions more efficient. 
Secondly, privacy risks emerge when agents access personal documents, as large language models may not fully adhere to privacy guidelines \cite{mireshghallah2023can}. Future work could focus on privacy-centric evaluations and explore new information access models to mitigate such risks.
Finally, an excessive number of AI-initiated requests can overwhelm users, hindering productivity. Building agents that can minimize human effort and prioritize urgent requests remains a challenge. 

\section*{Limitations}

\asyncfw{} consists of two tasks and is in one language (English). Future work could explore further expanding the domains and supported languages.
We make the simplifying assumption that an agent in our setup can engage only in dyadic conversations. Exploring more topologies such as group chats \cite{wu2023autogen} would bring-in additional challenges. 
We designed the domains and the experiment setup to study the effectiveness of the LM agents on a diverse set of information gathering behaviors. However, our analysis did not model all the possible factors in a real-world. Future work can explore additional factors such as turn-around speed and reliability of the response from a collaborator, how busy a person is, and various social dynamics that can be at play in organizations. 

\section*{Ethics Statement}
Allowing AI agents the capability to send messages to other users without fine-grained supervision presents a trade-off between saving user time and maintaining control. While autonomy can streamline workflows by eliminating the need for constant user confirmation, verifying key actions helps ensure accuracy and user oversight. While we studied the task in a sand-boxed environment, practitioners should carefully choose the degree of autonomy granted (for example, a more conservative approach would be to get user confirmation before every message that is sent). 

\section*{Acknowledgements}
We thank Jason Eisner and Hao Fang for thoughtful discussions. We thank Chris Kedzie, Patrick Xia, Justin Svegliato and Soham Dan for feedback on the paper.

%% file: appendix/10.4.approach.appendix.tex
\section{Additional details on approach}
\label{sec:appendix-approach}

\subsection{Action descriptions}
\label{sec:action_descriptions}

Listing \ref{ls:action-desc} shows the signatures and docstrings of Python functions corresponding to the set of allowed actions.

\subsection{Exemplars}
\label{sec:exemplars}

A fully annotated exemplar for question answering domain is provided in Listing \ref{lst:exemplar}, while a fully annotated exemplar for summarization domain is shown in Listing \ref{lst:exemplar2}.


\subsection{Action parsing failures}
We observed that parsing a predicted action into an allowed Python function and its parameters fails on a small fraction of cases. In such failure cases, we append a simple string `\# Failed to parse the predicted action. I will predict the next action again, following the allowed function signatures only' to the prompt, and decode the next action again. This process is followed upto a maximum of three times, post which the session is terminated.

\subsection{Overview of the prompt structure}
An overview of the prompt structure is available in Table \ref{tab:prompt}.

\begin{table}[h!]
\centering
\small 
\begin{tabular}{l}
\toprule
\texttt{Action Descriptions} \\
\midrule
\texttt{Exemplars}  \\
\midrule
\texttt{<Current Interaction>} \\
\verb|# {received-message}| \\
\verb|>>> {action-1}| \\
\verb|{observation-1}| \\
\verb|>>> {reflection-1}| \\
\verb|>>> {action-2}| \\
\ldots \\
\verb|>>> {turn-complete-action}| \\
\verb|# {received-message}| \\
\verb|>>> {action-1}| \\
\ldots \\
\hline
\end{tabular}
\caption{Overview of the prompt structure.}
\label{tab:prompt}
\end{table}

%% file: appendix/10.6.expsetup.appendix.tex
\section{Additional details on Experiment Setup}
\label{sec:appendix-eval}

\subsection{Match score}
\label{appendix-match-score}


\textbf{Prompt} outline used to compute Match score is shown in Listing \ref{lst:matchprompt}.

\lstset{
  basicstyle=\small\ttfamily,   
  columns=flexible,       
  breaklines=true,        
  frame=single
}

\begin{lstlisting}[breaklines=true, label={lst:matchprompt},caption={Prompt to compute Match score}]
# Task
- Agent serves as a digital assistant for {primary_user}. It can access different tools to send messages to other users. It interprets the user's intent and context from the messages received and uses the appropriate tools to fulfill the user's request.
- You will be given a dialogue between Agent and various human users. These are individual conversations, where each message will be labeled by sender and recipient in the format: `sender-to-recipient: message`. Your job is to rate Agent's response on a scale from 0 to 100, based on the rubric provided below.

## Scoring Scale
0. The response does not satisfy the criterion.
50. The response partially meets the criterion.
100. The response fully meets the criterion.

# Guidelines
- Format your output as follows using YAML:
```
criteria: <Copy of the criterion>
rationale: <Brief explanation of why you gave this score>
score: <Score between 0-100 based on how well Agent's response meets the criterion>
```
- Base your evaluation solely on the given criteria.
- If the criterion is clearly satisfied without any ambiguity, assign a full score of 100.
- Valid scores are 0, 50, or 100 only. 
- Formatting of the response shouldn't affect the score.
- Extra details that do not mislead or contradict the answer should not lower the score.

{examples}

# Conversation
Conversation Date: {conversation_date}
{conversation}

# Output
```
criteria: Agent should inform the original user that the answer to their question is {gold-answer}.
\end{lstlisting}

\textbf{Correlation with Human Rating:} 
One of the authors manually labeled 50 randomly selected outputs from \reactagent{} (with \texttt{gpt-4-turbo} as LLM) considering the same reference instructions and examples as in the prompt discussed above. We observe Cohen's Kappa score of $0.81$ between manual judgement ratings and Match score, suggesting a high agreement of the LLM-based Match metric with human judgement ratings.

\textbf{Stability:} 
We conducted an analysis where we rerun the Match scores in Table 1 three times, and the maximum change we observed in any value was 0.5 (Match is on a scale of 1-100), signifying very low instability issues. 
Additionally, we observed that switching the underlying LLM from \texttt{gpt-4-turbo} to \texttt{phi-3-medium} to compute Match scores resulted in the exact same ranking of the methods as in the results tables (Table \ref{tab:results-spider}), suggesting that relative performance of the methods under Match metric is stable with respect to the choice of the underlying LLM used to compute the metric.

\subsection{User Simulators}
\label{appendix-user-sim}

User simulator prompt, shown in Listings \ref{lst:sim-user}, consists of a basic set of instructions at the top, followed by five  examples of diverse situations a user can face (either as the initiating user, or as a teammate receiving a request). 
Each examples consists of a user description, the set of documents available to the user, and any conversation history so far.

\subsection{Qualitative Examples}
\label{appendix-qual}

Listings \ref{ls:qualshares} through \ref{ls:qualsummary} show randomly picked test examples from both the domains, demonstrating success as well as failure cases for \reactagent{}.

\subsection{Human Evaluation Study}
\label{appendix-humaneval}
Additional details about human participants: We recruited 5 participants, who each carried out 20 human-in-the-loop tasks. All the human participants are US graduates and well-versed with the English language. 
All participants are paid above the minimum wage requirements of the region.
Participants were given the same instructions and examples as in the simulated user prompt.  



%% file: appendix/10.0.data.appenidix.tex
\section{Additional details on datasets}
\label{sec:appendix-data}

\textsc{Spider} dataset is available under CC BY-SA 4.0 
 license.\footnote{\url{https://yale-lily.github.io/spider}}. \textsc{Multinews} dataset is available for research purposes.\footnote{\url{https://github.com/Alex-Fabbri/Multi-News/blob/master/LICENSE.txt}}

%% file: appendix/10.4.1.tex

\lstset{
  language=Python,                         
  basicstyle=\small\ttfamily,              
  keywordstyle=\color{keywordcolor}\bfseries,  
  stringstyle=\color{stringcolor},         
  commentstyle=\color{commentcolor}\itshape,  
  backgroundcolor=\color{bgcolor},         
  columns=flexible,                        
  breaklines=true,                         
  showspaces=false,                        
  showstringspaces=false,                  
  morekeywords={Enterprise,EnterpriseSearch,System,Reflection}, 
  morekeywords=[2]{resolve_person,send_message,thought,finish, search_relevant_people, search_documents, send_session_completed, resolve_person, resolve_primary_user},       
  keywordstyle=[2]\color{keywordcolor}\bfseries 
}
\onecolumn

\begin{longtable}{l}

\begin{lstlisting}[breaklines=true, caption={Action descriptions provided in the prompt, consisting of various function signatures and associated docstrings}, label={ls:action-desc}]
    
# You are a clever and helpful assistant helping a user. To accomplish the user request, you must use the following Python functions:

class System:

    # Functions
    def finish() -> None:
        """Call this function to indicate that the current turn is complete."""

class Enterprise:

    # Functions

    def send_message(user_id: str, content: str, title: str | None) -> None:
        """Send a message to a user."""

    def send_session_completed() -> None:
        """If the primary user indicates that they no longer need assistance, send a session completed message."""

    def resolve_primary_user() -> str:
        """Return the primary user details."""

    def resolve_person(name: str) -> str:
        """Find list of persons matching a given name and return details of the first match."""

class EnterpriseSearch:

    # Functions
    def search_documents(query: str) -> tuple[str, ...]:
        """Returns the list of relevant documents (including document content/records)"""

    def search_relevant_people(query: str) -> str:
        """Returns names of relevant person and any accompanying rationale"""

class Reflection:

    # Functions
    def thought(thought: str) -> None:
        """Assess the current state of the conversation and decide on the next action"""
\end{lstlisting}
\label{lst:action_descriptions}
\end{longtable}


\begin{lstlisting}[breaklines=true, caption={An annotated exemplar in the question answering domain, with received messages, tool calls, and observations.}, label={lst:exemplar}]
# Event: Alice Anastasiou says: I need first name and ages of the students who have a dog but do not have a cat as pet
>>> Reflection.thought("I will first resolve the primary user to get the right contact information")
>>> Enterprise.resolve_primary_user()
{"user_id":"alice@company.com", "email":"alice@company.com", "full_name": "Alice Anastasiou"}
>>> Reflection.thought("I will search user documents to see if the user request can be answered directly")
>>> EnterpriseSearch.search_documents(query="name and age of students with dogs but no cats as pets")
[
"Document:\nCollection name: pets 1\nTitle: pets\nRecord 1: {'pet id': 2001, 'pet type': 'cat', 'pet age': 3, 'weight': 12.0}\nRecord 2: {'pet id': 2002, 'pet type': 'dog', 'pet age': 2, 'weight': 13.4}\nRecord 3: {'pet id': 2003, 'pet type': 'dog', 'pet age': 1, 'weight': 9.3}\n"
]
>>> Reflection.thought("User documents have relevant information about pet ids and pet types. But there is not enough information to answer the query directly. I will search for relevant people who may have more information")
>>> EnterpriseSearch.search_relevant_people(query='name and age of students with dogs but no cats as pets')
{
  "Cassie Hicks": "User might have information about students and their pet IDs.",
  "Dewei Peng": "The user might have information about students, including their personal details, demographics, and academic information.",
  "Hannah Jarvis": "User might have information about students and their pet IDs."
}
>>> Reflection.thought("I can reach out to Hannah and Cassie who seem to have more information about pets owned by students. I might have to reach out to Dewei for the names and ages of students since Dewei seems to have personal details about students")
>>> Enterprise.send_message("alice@company.com", "Hi Alice, I couldn't find enough information in your documents to compile that directly. I can reach out to Hannah Jarvis, Cassie Hicks, and Dewei Peng who may have more information. Sounds good?")
>>> System.finish()

# Event: Alice Anastasiou says: Yes, that is fine.
>>> Reflection.thought("Both Hannah and Cassie have information about pets owned by students. Since I would need the list of all students who have a dog but not a cat as pet, I will contact both Hannah and Cassie. I will first resolve Hannah's contact information and then reach out to Hannah")
>>> Enterprise.resolve_person('Hannah')
{"user_id": "hannah", "email": "hannah@company.com", "full_name": "Hannah Jarvis"}
>>> Enterprise.send_message("hannah", "Hi Hannah, Can you please tell me the list of students and their pets?", title="Information needed about students")
>>> System.finish()

# Event: Hannah Jarvis says: Sure. I know that student with `student id`: 1001 has a pet with `pet id`: 2001 and student with `student id`: 1002 has a pet with `pet id`: 2002
>>> Reflection.thought("Hannah has provided the information about the pets owned by students. Since I need the list of all students who have a dog but not a cat as pet, I will reach out to Cassie in case Cassie knows more about the pets owned by students")
>>> Enterprise.resolve_person('Cassie')
{"user_id": "cassie", "email": "cassie@company.com", "full_name": "Cassie Hicks"}
>>> Enterprise.send_message("cassie", "Hi Cassie, Can you please tell me the list of students and their pets?", title="Information needed about students")
>>> System.finish()

# Event: Cassie Hicks says: Sure. 1) `student id`: 1002 has a pet with `pet id`: 2002; 2) `student id`: 1002 has a pet with `pet id`: 2003
>>> Reflection.thought("From information provided by Hannah and Cassie, and information in user documents, I can compile the list of students and their pet_ids as follows: `student id`: 1001 -> `pet id`: 2001, `student id`: 1002 -> `pet id`: 2002, `student id`: 1002 -> `pet id`: 2003. From Alice`s documents, I can get the type of animal for the pet ids: `pet id`: 2001 -> cat, `pet id`: 2002 -> dog, `pet id`: 2003 -> dog. I can conclude that the student with `student id`: 1002 has a dog but not a cat as a pet. I will reach out to Dewei for the first name and age of the student")
>>> Enterprise.resolve_person('Dewei')
{"user_id": "dewei", "email": "dewei@company.com", "full_name": "Dewei Peng"}
>>> Enterprise.send_message("dewei", "Hi Dewei, Can you please tell the name and age of student with `student id`: 1002?", title="Information needed about students")
>>> System.finish()

# Event: Dewei Peng says: Sorry, I do not have the information you are looking for. But you can try contacting Bhushan Magar who I think might have the information you are looking for.
>>> Reflection.thought("Dewei does not have the requested information on student names and ages. I will reach out to Bhushan Magar for the information. Since I haven't checked with Alice so far about reaching out to Bhushan, I will confirm it now")
>>> Enterprise.send_message("alice@company.com", "Hi Alice, I couldn't find enough information after reaching out to Dewei. I will now reach out to Bhushan Magar who may have more information. Sounds good?", title="Re: Clarification needed")
>>> System.finish()

# Event: Alice Anastasiou says: Sounds good, thanks for checking.
>>> Reflection.thought("Alice has confirmed that I can reach out to Bhushan. I will contact Bhushan requesting for the first name and age of the specified students")
>>> Enterprise.resolve_person('Bhushan')
{"user_id": "bhushan", "email": "bhushan@company.com", "full_name": "Bhushan Magar"}
>>> Enterprise.send_message("bhushan", "Hi Bhushan, Can you please tell the first name and age of `student id`: 1002?", title="Information needed about student")
>>> System.finish()

# Event: Bhushan Magar says: {"student id": 1002, "first name": "Tracy"}
>>> Reflection.thought("Bhushan has provided the first name of the student but not the age. I will reach out to Bhushan again for the age information. Since I already confirmed with Alice about reaching out to Bhushan, I will not confirm again")
>>> Enterprise.send_message("bhushan", "Thanks Bhushan! Can I also get the age information of `student id`: 1002?", title="Re: Information needed about student")
>>> System.finish()

# Event: Bhushan Magar says: Oh right, forgot about that. Age is 19.
>>> Reflection.thought("Bhushan has provided the name and age of the student. I now have all the information to answer Alice's query")
>>> Enterprise.send_message("alice@company.com", "Hi Alice, the first name and age of the student who has a dog but not a cat as a pet is:\n- First name: Tracy\n- Age: 19\nDo you need any more information?", title="Re: Information needed about student")
>>> System.finish()

# Event: Alice Anastasiou says: Thanks! That would be all.
>>> Enterprise.send_session_completed()
>>> Reflection.thought("Alice has confirmed that no more information is needed. I will close the session")
>>> System.finish()
\end{lstlisting} 


\begin{lstlisting}[breaklines=true, caption={An annotated exemplar in the summarization domain.}, label={lst:exemplar2}]
# Event: Alice Anastasiou says: I'm interested in compiling a summary on the topics: 'unemployment rate, jobs report, economy'.
>>> Reflection.thought("I will first resolve the primary user to get the right contact information")
>>> Enterprise.resolve_primary_user()
{"user_id": "alice", "email": "alice@company.com", "full_name": "Alice Anastasiou"}
>>> Reflection.thought("I will search user's documents and find relevant people who may have more information")
>>> EnterpriseSearch.search_documents(query="unemployment rate, jobs report, economy")
[]
>>> EnterpriseSearch.search_relevant_people(query="unemployment rate, jobs report, economy")
{
   "hannah": "Knows about the following topics:\nhiring, unemployment rate, job growth, economy, consumer spending\nGaithersburg, Maryland, jet crash, Montgomery County, FAA",
   "parker": "Knows about the following topics:\nShelly Sterling, LA Clippers sale, divorce\nGaithersburg, plane crash, Marie Gemmell, Michael Rosenberg, Montgomery County Air Park",
   "eden": "Knows about the following topics:\njobs, unemployment rate, economy"
}
>>> Reflection.thought("The current user has no documents. Hannah and Eden have information about the economy, so I should reach out to them.")
>>> Enterprise.send_message("alice", "Hi Alice, I couldn't find enough information in your documents to compile that information directly. I can reach out to Hannah and Eden who may have more information. Sound good?")
>>> System.finish()

# Event: Alice Anastasiou says: sure!
>>> Reflection.thought("I will reach out to Hannah to ask for information about the economy")
>>> Enterprise.resolve_person("hannah")
{"user_id": "hannah", "email": "hannah@company.com", "full_name": "Hannah Jarvis"}
>>> Enterprise.send_message("hannah", "Hi Hannah, do you have any information about the unemployment rate, jobs report, or economy?", title="Information about economy")
>>> System.finish()

# Event: Hannah Jarvis says: Sure, here's what I have: "employers pulled back sharply on hiring last month , a reminder that the u.s. economy may not be growing fast enough to sustain robust job growth . the unemployment rate dipped , but mostly because more americans stopped looking for work .     the labor department says the economy added 120,000 jobs in march , down from more than 200,000 in each of the previous three months .     the unemployment rate fell to 8.2 percent , th e lowest since january 2009. the rate dropped because fewer people searched for jobs . the official unemployment tally only includes those seeking work .     the economy has added 858,000 jobs since december _ the best four months of hiring in two years . but federal reserve chairman ben bernanke has cautioned that the current hiring pace is unlikely to continue without more consumer spending ."
>>> Reflection.thought("I will reach out to Eden to ask for information about the economy")
>>> Enterprise.resolve_person("Eden")
{"user_id": "eden", "email": "eden@company.com", "full_name": "Eden Berhe"}
>>> Enterprise.send_message("cassie", "Hi Eden, do you have any information about the unemployment rate, jobs report, or economy?", title="Information about economy")
>>> System.finish()

# Event: Eden Berhe says: Yes, I have an article with the following contents: "national archives     yes , it \u2019 s that time again , folks . it \u2019 s the first friday of the month , when for one ever-so-brief moment the interests of wall street , washington and main street are all aligned on one thing : jobs .     a fresh update on the u.s. employment situation for january hits the wires at 8 : 30 a.m. new york time offering one of the most important snapshots on how the economy fared during the previous month . expectations are for 203,000 new jobs to be created , according to economists polled by dow jones newswires , compared to 227,000 jobs added in february . the unemployment rate is expected to hold steady at 8.3 % .     here at marketbeat hq , we \u2019 ll be offering color commentary before and after the data crosses the wires . feel free to weigh-in yourself , via the comments section . and while you \u2019 re here , why don \u2019 t you sign up to  .     enjoy the show"
>>> Reflection.thought("I have enough information to compile the summary. I will send a message to Alice, with the summary enclosed in [[[ brackets ]]].")
>>> Enterprise.send_message("alice", "Hi Alice, here's the summary I put together: [[[ the unemployment rate dropped to 8.2 % last month , but the economy only added 120,000 jobs , when 203,000 new jobs had been predicted , according to today ' s jobs report . reaction on the wall street journal ' s marketbeat blog was swift : \" woah ! ! ! bad number . \" the unemployment rate , however , is better news ; it had been expected to hold steady at 8.3 % . but the ap notes that the dip is mostly due to more americans giving up on seeking employment ]]]", title="Economy summary")
>>> System.finish()

# Event: Alice Anastasiou says: Thanks! That would be all.
>>> Enterprise.send_session_completed()
>>> Reflection.thought("Alice has confirmed that no more information is needed. I will close the session")
>>> System.finish()
\end{lstlisting}

%% file: appendix/10.6.1.tex
\lstset{
  language=Markdown,                    
  basicstyle=\small\ttfamily,           
  backgroundcolor=\color{bgcolor},      
  breaklines=true,                      
  showspaces=false,                     
  showstringspaces=false,               
  breakatwhitespace=true,   
  columns=flexible
}

\onecolumn

\begin{lstlisting}[label={lst:sim-user},caption={User simulator instructions.}]
# Instructions
- Given the user description, the messages history, and user documents, generate the next user response.
- Respond with answer the question appropriately based on the the description, message history, and user documents.
- Respond with to-the-point statements. Format the text as needed.
- Respond by being a helpful colleague. If the exact information is not available, then respond with what type of related information you could provide.
- Respond with a single line message. Always end the message with a <eos>
- Respond with 'skip' only if agent's utterance is not a question

# Examples:

## Example
### User Description: Alice Anastasiou is an employee at Company. She is interested in knowing the names and ages of the students who have a dog but do not have a cat as pet which may be available in her documents or notes or Agent might have to reach out to other people. When Agent initates interaction, start with 'I needs names and ages of the students who have a dog but do not have a cat as pet'.
### Info: Today's date is Jan 1st, 2022, Monday.
### User Documents:
#### Document: 
Title: has pet
Record 1: {'student id': 1001, 'pet id': 2001}
Record 2: {'student id': 1002, 'pet id': 2002}
Record 3: {'student id': 1002, 'pet id': 2003}
### Messages History:
Agent: Hello, I'm here to help you as your Agent.
Alice: I needs names and ages of the students who have a dog but do not have a cat as pet
Agent: Hi Alice, I couldn't find enough information in your documents to compile that informatiom directly. Bhushan and Cassey might have more information. Who should I contact?
Alice: You can decide whom to contact.
Agent: Sounds good. I will reach out if I need any clarifications.
### Next Response:
Alice: skip <eos>

## Example
### User Description: Alice Anastasiou is an employee at Company. She is interested in knowing the names and ages of the students who have a dog but do not have a cat as pet which may be available in her documents or notes or Agent might have to reach out to other people. When Agent initates interaction, start with 'I needs names and ages of the students who have a dog but do not have a cat as pet'.
### Info: Today's date is Jan 1st, 2022, Monday.
### User Documents:
#### Document: 
Title: has pet
Record 1: {'student id': 1001, 'pet id': 2001}
Record 2: {'student id': 1002, 'pet id': 2002}
Record 3: {'student id': 1002, 'pet id': 2003}
### Messages History:
Agent: Hello, I'm here to help you as your Agent.
Alice: I needs names and ages of the students who have a dog but do not have a cat as pet
Agent: Hi Alice, I couldn't find enough information in your documents to compile that informatiom directly. I can reach out to Bhushan and Cassey who may have more information. Sounds good?
Alice: yes
Agent: Hi Alice, first name of students who have a dog but not a cat as pet:
- "first name": "Tracy"
Do you want to know their last names as well?
Alice: Thanks! I had also asked for their age.
Agent: Thanks for pointing that out. I found that Tracy's age is 19. Do you need any more information?
### Next Response:
Alice: No, that would be all. <eos>

## Example
### User Description: Bhushan Magar is an employee at Company. Bhushan will provide Agent with specific relevant information if it is available in his documents or notes
### User Documents:
#### Document:
Title: student
Record 1: {'student id': 1001, 'major': 600}
Record 2: {'student id': 1002, 'major': 600}
Record 3: {'student id': 1003, 'major': 600}
### Messages History:
Agent: Hi Bhushan, Can you please tell how many total students are there in the university?
### Next Response:
Bhushan: I know about major of 3 students. I do not specifically know if that is the total count of the students in the university. <eos>

## Example
### User Description: Cassie Hicks is an employee at Company. Cassie will provide Agent with specific relevant information if it is available in her documents or notes.
### Info: Today's date is Jan 1st, 2022, Monday.
### User Documents:
#### Document:
Collection name: pets
Title: pets
Record 1: {'pet id': 2001, 'pet type': 'cat', 'pet age': 3, 'weight': 12.0}
Record 2: {'pet id': 2002, 'pet type': 'dog', 'pet age': 2, 'weight': 13.4}
Record 3: {'pet id': 2003, 'pet type': 'dog', 'pet age': 1, 'weight': 9.3}
### Messages History:
Agent: Hi Cassie, can you please let me know which students have a dog as pet ? 
Cassie: I do not have that information. But I do know about pet ids and their types if you need that.
Agent: Thanks Cassie! Can you please let me know the type of animal are the following `pet ids`?
`pet id`: 2001
`pet id`: 2002
`pet id`: 2003
### Next Response:
Cassie: Sure. Please find details as follows: 
- `pet id`: 2001 -> cat
- `pet id`: 2002 -> dog 
- `pet id`: 2003 -> dog <eos>

## Example
### User Description: Bhushan Magar is an employee at Company. Bhushan will provide Agent with specific relevant information if it is available in his documents or notes
### User Documents:
#### Document:
Title: student
Record 1: {'student id': 1001, 'last name': 'Smith', 'first name': 'Linda', 'sex': 'F', 'major': 600, 'advisor': 1121, 'city code': 'BAL'}
Record 2: {'student id': 1002, 'last name': 'Kim', 'first name': 'Tracy', 'sex': 'F', 'major': 600, 'advisor': 7712, 'city code': 'HKG'}
Record 3: {'student id': 1003, 'last name': 'Jones', 'first name': 'Shiela', 'sex': 'F', 'major': 600, 'advisor': 7792, 'city code': 'WAS'}
### Messages History:
Agent: Hi Bhushan, Can you please tell the list of students who have a cat as pet ?
Bhushan: Sorry I do not have information about pet types. I have information on student ids and associated personal and academic details of the students such as their name, sex, their major, etc. Let me know if you need any of that information. 
Agent: Hi Bhushan, Can you please tell the first name and age of `student id`: 1002 ?
### Next Response:
Bhushan: First name of student id 1002 is 'Tracy'. I do not know their age. <eos>

# Task
## Example
    
\end{lstlisting}



\newpage

\input{appendix/qual-examples/game-injury}
\input{appendix/qual-examples/tracking-shares}
\input{appendix/qual-examples/customer-cards}
\input{appendix/qual-examples/college}

\input{appendix/qual-examples/summary}

%% file: appendix/qual-examples/game-injury.tex
\lstset{
  language=Python,                         
  basicstyle=\small\ttfamily,              
  keywordstyle=\color{keywordcolor}\bfseries,  
  stringstyle=\color{stringcolor},         
  commentstyle=\color{commentcolor}\itshape,  
  backgroundcolor=\color{bgcolor},         
  columns=flexible,                        
  breaklines=true,                         
  showspaces=false,                        
  showstringspaces=false,                  
  morekeywords={Enterprise,EnterpriseSearch,System,Reflection}, 
  morekeywords=[2]{resolve_person,send_message,thought,finish, search_relevant_people, search_documents, send_session_completed, resolve_person, resolve_primary_user},       
  keywordstyle=[2]\color{keywordcolor}\bfseries 
}

\begin{lstlisting}[breaklines=true, caption={Qualitative Example 1: An example trace, with \reactagent{} when using \texttt{gpt-4-turbo} as LLM. Agent is able to correctly handle `redirection' (Bhushan points to Sylvie) as well as `split document' (Eden and Kerstin each have information about different games.)}, label={ls:qualgame}]
# Alice Anastasiou says:  For each injury accident, find the date of the game and the name of the injured player in the game, and sort the results in descending order of game season. 
>>>  Reflection.thought("I will first resolve the primary user to get the right contact information")
>>>  Enterprise.resolve_primary_user()
{"email": "alice@company.com", "full_name": "Alice Anastasiou", "user_id": "alice@company.com"}
>>>  Reflection.thought("I will search user documents to see if the user request can be answered directly")
>>>  EnterpriseSearch.search_documents(query="injury accidents game date injured player")
('Document(url=\'game_injury/stadium\', title=\'stadium\', content=\'[{"id": 1, "name": "Cowboys Stadium", "home games": 8, "average attendance": 87047.0, "total attendance": 696377.0, "capacity percentage": 108.8}, {"id": 2, "name": "FedExField", "home games": 8, "average attendance": 83172.0, "total attendance": 665380.0, "capacity percentage": 90.7}, {"id": 3, "name": "New Meadowlands Stadium", "home games": 8, "average attendance": 79019.0, "total attendance": 632156.0, "capacity percentage": 95.8}, {"id": 4, "name": "New Meadowlands Stadium", "home games": 8, "average attendance": 78596.0, "total attendance": 628768.0, "capacity percentage": 95.3}, {"id": 5, "name": "Invesco Field at Mile High", "home games": 8, "average attendance": 74908.0, "total attendance": 599264.0, "capacity percentage": 98.4}, {"id": 6, "name": "Bank of America Stadium", "home games": 8, "average attendance": 72620.0, "total attendance": 580965.0, "capacity percentage": 98.4}, {"id": 7, "name": "M&T Bank Stadium", "home games": 8, "average attendance": 71227.0, "total attendance": 569817.0, "capacity percentage": 100.3}, {"id": 8, "name": "Reliant Stadium", "home games": 8, "average attendance": 71080.0, "total attendance": 568643.0, "capacity percentage": 100.0}, {"id": 9, "name": "Lambeau Field", "home games": 8, "average attendance": 70795.0, "total attendance": 566362.0, "capacity percentage": 97.1}, {"id": 10, "name": "Louisiana Superdome", "home games": 8, "average attendance": 70038.0, "total attendance": 560304.0, "capacity percentage": 96.0}, {"id": 11, "name": "Lincoln Financial Field", "home games": 8, "average attendance": 69144.0, "total attendance": 553152.0, "capacity percentage": 102.3}, {"id": 12, "name": "LP Field", "home games": 8, "average attendance": 69143.0, "total attendance": 553144.0, "capacity percentage": 100.0}, {"id": 13, "name": "Gillette Stadium", "home games": 8, "average attendance": 68756.0, "total attendance": 550048.0, "capacity percentage": 100.0}]\')',)
>>>  Reflection.thought("The document found does not contain information about injury accidents, game dates, or injured players. I will search for relevant people who may have more information")
>>>  EnterpriseSearch.search_relevant_people(query="injury accidents game date injured player")
{"kerstin": " The user might have information about games, including the stadium, season, date, teams, score, and competition.", "eden": " The user might have information about games, including the stadium, season, date, teams, score, and competition.", "bhushan": " The user might have information about injury accidents, including the game id, player, injury, number of matches, and source."}
>>>  Reflection.thought("I can reach out to Kerstin, Eden, and Bhushan who seem to have more information about games and injury accidents. I will confirm with Alice if that is okay")
>>>  Enterprise.send_message("alice@company.com", "Hi Alice, I couldn't find enough information in your documents to compile that directly. I can reach out to Kerstin, Eden, and Bhushan who may have more information. Sounds good?")
>>>  System.finish()

# Alice Anastasiou says:  Yes, please reach out to them. 
>>>  Reflection.thought("Alice has confirmed that I can reach out to Kerstin, Eden, and Bhushan. I will start by reaching out to Bhushan for information about injury accidents")
>>>  Enterprise.resolve_person("Bhushan")
{"email": "bhushan@company.com", "full_name": "Bhushan Magar", "user_id": "bhushan"}
>>>  Enterprise.send_message("bhushan@company.com", "Hi Bhushan, Can you please provide information about injury accidents, including the game id and the name of the injured player?", title="Information needed about injury accidents")
>>>  System.finish()

# Bhushan Magar says:  I do not have that information. You can try contacting Sylvie Rocher for the same. 
>>>  Reflection.thought("Bhushan does not have the requested information but suggested contacting Sylvie Rocher. I will reach out to Sylvie for the information")
>>>  Enterprise.resolve_person("Sylvie")
{"email": "sylvie@company.com", "full_name": "Sylvie Rocher", "user_id": "sylvie"}
>>>  Enterprise.send_message("sylvie@company.com", "Hi Sylvie, Can you please provide information about injury accidents, including the game id and the name of the injured player?", title="Information needed about injury accidents")
>>>  System.finish()

# Sylvie Rocher says:  I have information on game injuries including game id, player name, type of injury, number of matches missed, and the source. 
>>>  Reflection.thought("Sylvie has the information about game injuries. I will ask Sylvie for the game id and the name of the injured player for each injury accident")
>>>  Enterprise.send_message("sylvie@company.com", "Hi Sylvie, Can you please provide the game id and the name of the injured player for each injury accident?", title="Re: Information needed about injury accidents")
>>>  System.finish()

# Sylvie Rocher says:  Here are the game ids and names of injured players for each injury accident:
- Game id: 1, Player: Davide Santon
- Game id: 2, Player: Maicon
- Game id: 3, Player: Dejan Stankovic
- Game id: 4, Player: Wesley Sneijder
- Game id: 5, Player: Diego Milito
- Game id: 5, Player: Javier Zanetti
- Game id: 6, Player: Walter Samuel
- Game id: 6, Player: Thiago Motta
- Game id: 6, Player: Goran Pandev
- Game id: 8, Player: Ivan Cordoba
- Game id: 9, Player: Esteban Cambiasso
- Game id: 10, Player: Dejan Stankovic (2)
- Game id: 10, Player: Wesley Sneijder (2)
- Game id: 11, Player: Esteban Cambiasso (2)
- Game id: 12, Player: Thiago Motta (2) 
>>>  Reflection.thought("Sylvie has provided the game ids and names of injured players. I will now reach out to Kerstin and Eden for the game dates and seasons")
>>>  Enterprise.resolve_person("Kerstin")
{"email": "kerstin@company.com", "full_name": "Kerstin Mark", "user_id": "kerstin"}
>>>  Enterprise.send_message("kerstin@company.com", "Hi Kerstin, Can you please provide the date of the game and the season for the following game ids: 1, 2, 3, 4, 5, 6, 8, 9, 10, 11, 12?", title="Information needed about games")
>>>  System.finish()

# Kerstin Mark says:  Here are the details for the game ids you requested:
- Game id 1: Date - 18 May 2007, Season - 2007
- Game id 2: Date - 22 September 2007, Season - 2007
- Game id 3: Date - 9 December 2007, Season - 2007
- Game id 4: Date - 10 July 2008, Season - 2008
- Game id 5: Date - 16 August 2008, Season - 2008
- Game id 6: Date - 31 October 2008, Season - 2008
- Game id 8: Date - 14 October 2009, Season - 2009
- Game id 9: Date - 8 August 2009, Season - 2009
- Game id 10: Date - 14 March 2010, Season - 2010
I do not have information for game ids 11 and 12. 
>>>  Reflection.thought("Kerstin has provided the game dates and seasons for most game ids but not for 11 and 12. I will reach out to Eden for the missing information")
>>>  Enterprise.resolve_person("Eden")
{"email": "eden@company.com", "full_name": "Eden Berhe", "user_id": "eden"}
>>>  Enterprise.send_message("eden@company.com", "Hi Eden, Can you please provide the date of the game and the season for the following game ids: 11, 12?", title="Information needed about games")
>>>  System.finish()

# Eden Berhe says:  The date for game id 11 is 31 October 2010 and the season is 2010. The date for game id 12 is 7 July 2011 and the season is 2011. 
>>>  Reflection.thought("Eden has provided the game dates and seasons for game ids 11 and 12. I now have all the information needed to answer the query. I will compile the information and send it to Alice")
>>>  Enterprise.send_message("alice@company.com", "Hi Alice, here are the dates of the games and the names of the injured players for each injury accident, sorted in descending order of game season:\n\n- Season 2011: Game date - 7 July 2011, Player - Thiago Motta (2)\n- Season 2010: Game date - 31 October 2010, Player - Esteban Cambiasso (2)\n- Season 2010: Game date - 14 March 2010, Players - Dejan Stankovic (2), Wesley Sneijder (2)\n- Season 2009: Game date - 14 October 2009, Player - Ivan Cordoba\n- Season 2009: Game date - 8 August 2009, Player - Esteban Cambiasso\n- Season 2008: Game date - 31 October 2008, Players - Walter Samuel, Thiago Motta, Goran Pandev\n- Season 2008: Game date - 16 August 2008, Players - Diego Milito, Javier Zanetti\n- Season 2008: Game date - 10 July 2008, Player - Wesley Sneijder\n- Season 2007: Game date - 9 December 2007, Player - Dejan Stankovic\n- Season 2007: Game date - 22 September 2007, Player - Maicon\n- Season 2007: Game date - 18 May 2007, Player - Davide Santon\n\nPlease let me know if you need any further information.", title="Injury accidents sorted by game season")
>>>  System.finish()

# Alice Anastasiou says:   No, that would be all. Thank you! 
>>>  Reflection.thought("Alice has confirmed that no more information is needed. I will close the session")
>>>  Enterprise.send_session_completed()
>>>  System.finish()

\end{lstlisting}

%% file: appendix/qual-examples/tracking-shares.tex
\begin{lstlisting}[breaklines=true, caption={Qualitative Example 2: An example trace, with \reactagent{} when using \texttt{gpt-4-turbo} as LLM. Agent, after searching in Alice's documents and reaching out to seeingly relevant people, correctly concludes that there is simply not enough information to answer Alice's request (although it could have potentially arrived at the conclusion with fewer messages). Match score = 100.}, label={ls:qualshares}]
# Alice Anastasiou says:  Show the average share count of transactions for different investors. 
>>>  Reflection.thought("I will first resolve the primary user to get the right contact information")
>>>  Enterprise.resolve_primary_user()
{"email": "alice@company.com", "full_name": "Alice Anastasiou", "user_id": "alice@company.com"}
>>>  Reflection.thought("I will search user documents to see if the user request can be answered directly")
>>>  EnterpriseSearch.search_documents(query="average share count of transactions for different investors")
('Document(url=\'tracking_share_transactions/investors\', title=\'investors\', content=\'[{"investor id": 1, "investor details": "z"}, {"investor id": 2, "investor details": "z"}, {"investor id": 3, "investor details": "d"}, {"investor id": 4, "investor details": "d"}, {"investor id": 5, "investor details": "b"}, {"investor id": 6, "investor details": "k"}, {"investor id": 7, "investor details": "l"}, {"investor id": 8, "investor details": "t"}, {"investor id": 9, "investor details": "y"}, {"investor id": 10, "investor details": "r"}, {"investor id": 11, "investor details": "q"}, {"investor id": 12, "investor details": "c"}, {"investor id": 13, "investor details": "o"}, {"investor id": 14, "investor details": "w"}, {"investor id": 15, "investor details": "i"}, {"investor id": 16, "investor details": "y"}, {"investor id": 17, "investor details": "k"}, {"investor id": 18, "investor details": "w"}, {"investor id": 19, "investor details": "l"}, {"investor id": 20, "investor details": "j"}]\')',)
>>>  Reflection.thought("The document found does not contain information about share count of transactions. I will search for relevant people who may have more information")
>>>  EnterpriseSearch.search_relevant_people(query="average share count of transactions for different investors")
{"tulga": " The user might have information about lots, including the lot id, investor id, and lot details.", "bhushan": " The user might have information about sales transactions and their details.", "farshid": " The user might have information about purchases, including transaction ids and details.", "irena": " The user might have information about transactions and their associated lot ids.", "cassie": " The user might have information about reference transaction types, including the transaction type code and description."}
>>>  Reflection.thought("I can reach out to Tulga, Bhushan, Farshid, Irena, and Cassie who seem to have more information about transactions. I will confirm with Alice if that is okay")
>>>  Enterprise.send_message("alice@company.com", "Hi Alice, I couldn't find enough information in your documents to compile that directly. I can reach out to Tulga, Bhushan, Farshid, Irena, and Cassie who may have more information. Sounds good?")
>>>  System.finish()

# Alice Anastasiou says:  Yes, please reach out to them. 
>>>  Reflection.thought("Alice has confirmed that I can reach out to Tulga, Bhushan, Farshid, Irena, and Cassie. I will start by reaching out to Tulga for information about lots and investor ids")
>>>  Enterprise.resolve_person("Tulga")
{"email": "tulga@company.com", "full_name": "Tulga Bat-Erdene", "user_id": "tulga"}
>>>  Enterprise.send_message("tulga@company.com", "Hi Tulga, Can you please provide information about lots, including the lot id, investor id, and lot details?", title="Information needed about lots")
>>>  System.finish()

# Tulga Bat-Erdene says:  I have information on 15 lots with their investor ids and lot details. 
>>>  Reflection.thought("Tulga has information on lots but did not provide specific details. I will ask Tulga for the specific lot details and investor ids")
>>>  Enterprise.send_message("tulga@company.com", "Thanks, Tulga. Can you please share the specific lot details and investor ids for those 15 lots?", title="Re: Information needed about lots")
>>>  System.finish()

# Tulga Bat-Erdene says:  Here are the details for the 15 lots:
- Lot 1: Investor ID 13, Details 'r'
- Lot 2: Investor ID 16, Details 'z'
- Lot 3: Investor ID 10, Details 's'
- Lot 4: Investor ID 19, Details 's'
- Lot 5: Investor ID 6, Details 'q'
- Lot 6: Investor ID 20, Details 'd'
- Lot 7: Investor ID 7, Details 'm'
- Lot 8: Investor ID 7, Details 'h'
- Lot 9: Investor ID 20, Details 'z'
- Lot 10: Investor ID 9, Details 'x'
- Lot 11: Investor ID 1, Details 'd'
- Lot 12: Investor ID 19, Details 'm'
- Lot 13: Investor ID 7, Details 'z'
- Lot 14: Investor ID 6, Details 'd'
- Lot 15: Investor ID 1, Details 'h' 
>>>  Reflection.thought("Tulga has provided lot details and investor ids. I will now reach out to Bhushan for information about sales transactions")
>>>  Enterprise.resolve_person("Bhushan")
{"email": "bhushan@company.com", "full_name": "Bhushan Magar", "user_id": "bhushan"}
>>>  Enterprise.send_message("bhushan@company.com", "Hi Bhushan, Can you please provide information about sales transactions and their details?", title="Information needed about sales transactions")
>>>  System.finish()

# Bhushan Magar says:  I have information on 15 sales transactions and their details. 
>>>  Reflection.thought("Bhushan has information on sales transactions but did not provide specific details. I will ask Bhushan for the specific sales transaction details")
>>>  Enterprise.send_message("bhushan@company.com", "Thanks, Bhushan. Can you please share the specific details for those 15 sales transactions?", title="Re: Information needed about sales transactions")
>>>  System.finish()

# Bhushan Magar says:  Sure, here are the details for the 15 sales transactions: 
- 'sales transaction id': 1, 'sales details': 'x'
- 'sales transaction id': 2, 'sales details': 'o'
- 'sales transaction id': 3, 'sales details': 'a'
- 'sales transaction id': 4, 'sales details': 'f'
- 'sales transaction id': 5, 'sales details': 'y'
- 'sales transaction id': 6, 'sales details': 'x'
- 'sales transaction id': 7, 'sales details': 'p'
- 'sales transaction id': 8, 'sales details': 'e'
- 'sales transaction id': 9, 'sales details': 'p'
- 'sales transaction id': 10, 'sales details': 's'
- 'sales transaction id': 11, 'sales details': 's'
- 'sales transaction id': 12, 'sales details': 't'
- 'sales transaction id': 13, 'sales details': 'p'
- 'sales transaction id': 14, 'sales details': 'n'
- 'sales transaction id': 15, 'sales details': 'e' 
>>>  Reflection.thought("Bhushan has provided sales transaction details. I will now reach out to Farshid for information about purchases")
>>>  Enterprise.resolve_person("Farshid")
{"email": "farshid@company.com", "full_name": "Farshid Kamangar", "user_id": "farshid"}
>>>  Enterprise.send_message("farshid@company.com", "Hi Farshid, Can you please provide information about purchases, including transaction ids and details?", title="Information needed about purchases")
>>>  System.finish()

# Farshid Kamangar says:  Sure, here are the purchase transaction ids and their details:
- `purchase transaction id`: 1 -> 'c'
- `purchase transaction id`: 2 -> 'y'
- `purchase transaction id`: 3 -> 'i'
- `purchase transaction id`: 4 -> 'x'
- `purchase transaction id`: 5 -> 'y'
- `purchase transaction id`: 6 -> 'a'
- `purchase transaction id`: 7 -> 'r'
- `purchase transaction id`: 8 -> 'a'
- `purchase transaction id`: 9 -> 'r'
- `purchase transaction id`: 10 -> 'l'
- `purchase transaction id`: 11 -> 'z'
- `purchase transaction id`: 12 -> 'h'
- `purchase transaction id`: 13 -> 't'
- `purchase transaction id`: 14 -> 'o'
- `purchase transaction id`: 15 -> 'x' 
>>>  Reflection.thought("Farshid has provided purchase transaction details. I will now reach out to Irena for information about transactions and their associated lot ids")
>>>  Enterprise.resolve_person("Irena")
{"email": "irena@company.com", "full_name": "Irena Jovanovic", "user_id": "irena"}
>>>  Enterprise.send_message("irena@company.com", "Hi Irena, Can you please provide information about transactions and their associated lot ids?", title="Information needed about transactions")
>>>  System.finish()

# Irena Jovanovic says:  Sure, here are the details:
- `transaction id`: 3 -> `lot id`: 11, 8, 14, 4
- `transaction id`: 2 -> `lot id`: 11, 1
- `transaction id`: 12 -> `lot id`: 10, 15
- `transaction id`: 15 -> `lot id`: 10
- `transaction id`: 10 -> `lot id`: 10
- `transaction id`: 1 -> `lot id`: 1, 14
- `transaction id`: 14 -> `lot id`: 9
- `transaction id`: 7 -> `lot id`: 1
- `transaction id`: 6 -> `lot id`: 3 
>>>  Reflection.thought("Irena has provided transaction ids and their associated lot ids. I will now reach out to Cassie for information about reference transaction types")
>>>  Enterprise.resolve_person("Cassie")
{"email": "cassie@company.com", "full_name": "Cassie Hicks", "user_id": "cassie"}
>>>  Enterprise.send_message("cassie@company.com", "Hi Cassie, Can you please provide information about reference transaction types, including the transaction type code and description?", title="Information needed about transaction types")
>>>  System.finish()

# Cassie Hicks says:  I do not have that information, but you can try contacting Juan Quispe for it. 
>>>  Reflection.thought("Cassie does not have the requested information but suggested contacting Juan Quispe. I will reach out to Juan for the information about reference transaction types")
>>>  Enterprise.resolve_person("Juan")
{"email": "juan@company.com", "full_name": "Juan Quispe", "user_id": "juan"}
>>>  Enterprise.send_message("juan@company.com", "Hi Juan, Can you please provide information about reference transaction types, including the transaction type code and description?", title="Information needed about transaction types")
>>>  System.finish()

# Juan Quispe says:  Sure, here are the reference transaction types:
- `transaction type code`: SALE -> `transaction type description`: Sale
- `transaction type code`: PUR -> `transaction type description`: Purchase 
>>>  Reflection.thought("Juan has provided the reference transaction types. I have now gathered information from Tulga, Bhushan, Farshid, Irena, and Juan. However, I do not have the specific share count of transactions to calculate the average for different investors. I will inform Alice that I could not find the specific share count information.")
>>>  Enterprise.send_message("alice@company.com", "Hi Alice, I have gathered information from Tulga, Bhushan, Farshid, Irena, and Juan. However, I do not have the specific share count of transactions to calculate the average for different investors. I'm sorry, but I couldn't find the specific share count information you requested.", title="Information about average share count")
>>>  Enterprise.send_session_completed()
>>>  System.finish()
\end{lstlisting}

%% file: appendix/qual-examples/customer-cards.tex
\begin{lstlisting}[breaklines=true, caption={Qualitative Example 3: An example trace, with \reactagent{} when using \texttt{gpt-4-turbo} as LLM. Agent decided to not search for relevant people, and wasn't able to reach to the correct outcome. Match score = 0.}, label={ls:qualcustomer}]
# Alice Anastasiou says:  What are the different card-type codes, and how many different customers hold each type? 
>>>  Reflection.thought("I will first resolve the primary user to get the right contact information")
>>>  Enterprise.resolve_primary_user()
{"email": "alice@company.com", "full_name": "Alice Anastasiou", "user_id": "alice@company.com"}
>>>  Reflection.thought("I will search user documents to see if the user request can be answered directly")
>>>  EnterpriseSearch.search_documents("different card type codes and number of customers holding each type")
('Document(url=\'customers_card_transactions/accounts\', title=\'accounts\', content=\'[{"account id": 1, "customer id": 6, "account name": "338", "other account details": "Regular"}, {"account id": 2, "customer id": 14, "account name": "562", "other account details": "VIP"}, {"account id": 3, "customer id": 9, "account name": "162", "other account details": "VIP"}, {"account id": 4, "customer id": 12, "account name": "038", "other account details": "Regular"}, {"account id": 5, "customer id": 13, "account name": "858", "other account details": "Regular"}, {"account id": 6, "customer id": 12, "account name": "262", "other account details": "Regular"}, {"account id": 7, "customer id": 9, "account name": "621", "other account details": "Regular"}, {"account id": 8, "customer id": 8, "account name": "381", "other account details": "Regular"}, {"account id": 9, "customer id": 9, "account name": "546", "other account details": "Regular"}, {"account id": 10, "customer id": 4, "account name": "767", "other account details": "VIP"}, {"account id": 11, "customer id": 4, "account name": "022", "other account details": "VIP"}, {"account id": 12, "customer id": 2, "account name": "866", "other account details": "VIP"}, {"account id": 13, "customer id": 10, "account name": "234", "other account details": "VIP"}, {"account id": 14, "customer id": 9, "account name": "557", "other account details": "VIP"}, {"account id": 15, "customer id": 4, "account name": "725", "other account details": "VIP"}]\')',)
>>>  Reflection.thought("The document contains account details but does not specify card type codes or the number of customers holding each type. I will inform Alice that the information is not available in the documents.")
>>>  Enterprise.send_message("alice@company.com", "Hi Alice, I searched through your documents but couldn't find specific information on different card type codes and the number of customers holding each type. The accounts document only contains account ids, customer ids, account names, and other account details, but no information on card types or customer counts for each type.", title="Card Type Codes Information")
>>>  System.finish()
\end{lstlisting}

%% file: appendix/qual-examples/college.tex
\begin{lstlisting}[breaklines=true, caption={Qualitative Example 4: An example trace, with \reactagent{} when using \texttt{gpt-4-turbo} as LLM. Agent found the correct people to reach out, but didn't frame/reframe the correct questions. Ideally, agent should have asked Cassie about course enrollments for course-ids, and then ask Niks to get names for those course-ids; or attempt to re-frame the questions once they responded back. Match score = 0.  )}, label={ls:qualcollege}]
# Alice Anastasiou says:  Give the names of the courses with at least five enrollments. 
>>>  Reflection.thought("I will first resolve the primary user to get the right contact information")
>>>  Enterprise.resolve_primary_user()
{"email": "alice@company.com", "full_name": "Alice Anastasiou", "user_id": "alice@company.com"}
>>>  Reflection.thought("I will search user documents to see if the user request can be answered directly")
>>>  EnterpriseSearch.search_documents("courses with at least five enrollments")
('Document(url=\'college_3/student\', title=\'student\', content=\'[{"student id": 1001, "last name": "Smith", "first name": "Linda", "age": 18, "sex": "F", "major": 600, "advisor": 1121, "city code": "BAL"}, {"student id": 1002, "last name": "Kim", "first name": "Tracy", "age": 19, "sex": "F", "major": 600, "advisor": 7712, "city code": "HKG"}, {"student id": 1003, "last name": "Jones", "first name": "Shiela", "age": 21, "sex": "F", "major": 600, "advisor": 7792, "city code": "WAS"}, {"student id": 1004, "last name": "Kumar", "first name": "Dinesh", "age": 20, "sex": "M", "major": 600, "advisor": 8423, "city code": "CHI"}, {"student id": 1005, "last name": "Gompers", "first name": "Paul", "age": 26, "sex": "M", "major": 600, "advisor": 1121, "city code": "YYZ"}, {"student id": 1006, "last name": "Schultz", "first name": "Andy", "age": 18, "sex": "M", "major": 600, "advisor": 1148, "city code": "BAL"}, {"student id": 1007, "last name": "Apap", "first name": "Lisa", "age": 18, "sex": "F", "major": 600, "advisor": 8918, "city code": "PIT"}, {"student id": 1008, "last name": "Nelson", "first name": "Jandy", "age": 20, "sex": "F", "major": 600, "advisor": 9172, "city code": "BAL"}, {"student id": 1009, "last name": "Tai", "first name": "Eric", "age": 19, "sex": "M", "major": 600, "advisor": 2192, "city code": "YYZ"}, {"student id": 1010, "last name": "Lee", "first name": "Derek", "age": 17, "sex": "M", "major": 600, "advisor": 2192, "city code": "HOU"}]\')',)
>>>  Reflection.thought("The document searched does not contain information about course enrollments. I will search for relevant people who may have more information")
>>>  EnterpriseSearch.search_relevant_people("courses with at least five enrollments")
{"eden": " The user might have information about faculty members, including their personal details, rank, contact information, and office location.", "dewei": " The user might have information about departments, including their names, divisions, rooms, buildings, and phone numbers.", "niks": " The user might have information about courses, including the course name, credits, instructor, schedule, and department number.", "parker": " The user might have information about students and their department numbers.", "cassie": " The user might have information about students and their course enrollments, including grades.", "tulga": " The user might have information about grade conversion, including letter grades and grade points."}
>>>  Reflection.thought("I can reach out to Cassie who seems to have information about course enrollments. I will confirm with Alice if it's okay to reach out to Cassie.")
>>>  Enterprise.send_message("alice@company.com", "Hi Alice, I couldn't find enough information in your documents to compile that directly. I can reach out to Cassie who may have more information about course enrollments. Sounds good?")
>>>  System.finish()

# Alice Anastasiou says:  Yes, please do that. 
>>>  Reflection.thought("Alice has confirmed that I can reach out to Cassie for information about course enrollments.")
>>>  Enterprise.resolve_person("Cassie")
{"email": "cassie@company.com", "full_name": "Cassie Hicks", "user_id": "cassie"}
>>>  Enterprise.send_message("cassie@company.com", "Hi Cassie, Can you please provide the names of the courses with at least five enrollments?", title="Information needed about courses")
>>>  System.finish()

# Cassie Hicks says:  I'm sorry, I don't have the names of the courses, only the course ids and the grades of the students enrolled in them. 
>>>  Reflection.thought("Cassie does not have the names of the courses, only the course ids and grades. I will reach out to Niks who might have information about courses, including the course name.")
>>>  Enterprise.resolve_person("Niks")
{"email": "niks@company.com", "full_name": "Niks Dzenis", "user_id": "niks"}
>>>  Enterprise.send_message("niks@company.com", "Hi Niks, Can you please provide the names of the courses with at least five enrollments?", title="Information needed about courses")
>>>  System.finish()

# Niks Dzenis says:  I'm sorry, I don't have information on course enrollments, but I can provide details on course names, credits, instructors, and schedules if that helps. 
>>>  Reflection.thought("Niks does not have information on course enrollments. I will inform Alice that I could not find the names of the courses with at least five enrollments.")
>>>  Enterprise.send_message("alice@company.com", "Hi Alice, I reached out to Cassie and Niks, but unfortunately, I couldn't find the names of the courses with at least five enrollments. Cassie only has course ids and grades, and Niks does not have information on course enrollments. I'm sorry I couldn't fulfill your request. Is there anything else I can assist you with?")
>>>  System.finish()

# Alice Anastasiou says:  No, that's all for now. Thank you for trying. 
>>>  Reflection.thought("Alice has confirmed that no more information is needed. I will close the session")
>>>  Enterprise.send_session_completed()
>>>  System.finish()  # Closing the session as requested by the user.
\end{lstlisting}

%% file: appendix/qual-examples/summary.tex
\begin{lstlisting}[breaklines=true, caption={Qualitative Example 5: An example trace in the summarization domain, with \reactagent{} when using \texttt{gpt-4-turbo} as LLM.},label={ls:qualsummary}]
# Alice Anastasiou says: I'm interested in compiling a summary on the topic: governor, race, gop, obamacare, medicaid. Can you help me? 
>>>  Reflection.thought("I will first resolve the primary user to get the right contact information")
>>>  Enterprise.resolve_primary_user()
{"email": "alice@company.com", "full_name": "Alice Anastasiou", "user_id": "alice"}
>>>  Reflection.thought("I will search user's documents and find relevant people who may have more information")
>>>  EnterpriseSearch.search_documents(query="governor, race, gop, obamacare, medicaid")
>>>  EnterpriseSearch.search_relevant_people(query="governor, race, gop, obamacare, medicaid")
{"parker": "Knows about the following topics:\ngovernor elections, GOP, health care, Obama, Romney", "bhushan": "Knows about the following topics:\ngovernor elections, GOP, health care, Obama, Romney\nSiskiyou County, California, secession, state of Jefferson, rural representation", "farshid": "Knows about the following topics:\nSiskiyou County, secession, Jefferson, California, rural fire prevention fee", "eden": "Knows about the following topics:\nFacebook, gay rights, John Snow kiss-in, Richard Metzger, Paul Shetler\nFacebook, same-sex kissing, photo removal, protest", "hannah": "Knows about the following topics:\n", "dewei": "Knows about the following topics:\n", "cassie": "Knows about the following topics:\n"}
>>>  Reflection.thought("Parker and Bhushan have information about the governor race and GOP, so I should reach out to them.")
>>>  Enterprise.send_message("alice", "Hi Alice, I can reach out to Parker and Bhushan to produce a summary. Is that OK?")
>>>  System.finish()

# Alice Anastasiou says: Yes, please do that. 
>>>  Reflection.thought("I will reach out to Parker to ask for information about the governor race and GOP")
>>>  Enterprise.resolve_person("Parker")
{"email": "parker@company.com", "full_name": "Parker McLean", "user_id": "parker"}
>>>  Enterprise.send_message("parker", "Hi Parker, do you have any information about the governor race, GOP, Obamacare, or Medicaid?", title="Information about governor race and GOP")
>>>  System.finish()

# Parker McLean says: Yes, here's a relevant document I found: "gop eyes gains as voters in 11 states pick governors     enlarge this image toggle caption jim cole/ap jim cole/ap     voters in 11 states will pick their governors tonight , and republicans appear on track to increase [...] "
>>>  Reflection.thought("I will reach out to Bhushan to ask for information about the governor race and GOP")
>>>  Enterprise.resolve_person("Bhushan")
{"email": "bhushan@company.com", "full_name": "Bhushan Magar", "user_id": "bhushan"}
>>>  Enterprise.send_message("bhushan", "Hi Bhushan, do you have any information about the governor race, GOP, Obamacare, or Medicaid?", title="Information about governor race and GOP")
>>>  System.finish()

# Bhushan Magar says: Yes, here's a relevant document I found: "the real clear politics average for the race has mccrory maintaining a 14.3 percentage point lead .     montana : popular democratic gov . brian schweitzer - he won his last election with 65 percent of the vote - has [...]" 
>>>  Reflection.thought("I have enough information to compile the summary. I will send a message to Alice, with the summary enclosed in [[[ brackets ]]].")
>>>  Enterprise.send_message("alice", "Hi Alice, here's the summary I put together: [[[Voters in 11 states are choosing their governors, with Republicans poised to increase their numbers, potentially controlling over two-thirds of the nation's top state offices. The GOP is expected to take North Carolina and retain seats in Utah, North Dakota, and Indiana. Democrats are likely to hold West Virginia and Missouri, with competitive races in Montana, New Hampshire, and Washington. The governor's role is crucial in healthcare, as they can opt out of Medicaid expansion under the Affordable Care Act. A Romney win would empower Republican governors, especially in states where Obama's policies, like high-speed rail funding, were previously rejected.]]]", title="Governor race and GOP summary")

# Alice Anastasiou says: Thank you, this is exactly what I needed. 
>>>  Reflection.thought("Alice has confirmed that no more information is needed. I will close the session")
>>>  Enterprise.send_session_completed()
>>>  System.finish()
\end{lstlisting}